# Comparing roughness maps generated by five roughness descriptors for LiDAR-derived digital elevation models

**Lei Fan [1] and Yang Zhao [2]**

[1]  Department of Civil Engineering, Xi'an Jiaotong-Liverpool University, Suzhou, China
[2]  School of Intelligent Manufacturing and Smart Transportation, Suzhou City University, Suzhou, China

**Abstract:** Terrain surface roughness, often described abstractly, poses challenges in quantitative characterisation with various descriptors found in the literature. This study compares five commonly used roughness descriptors, exploring correlations among their quantified terrain surface roughness maps across three terrains with distinct spatial variations. Additionally, the study investigates the impacts of spatial scales and interpolation methods on these correlations. Dense point cloud data obtained through Light Detection and Ranging technique are used in this study. The findings highlight both global pattern similarities and local pattern distinctions in the derived roughness maps, emphasizing the significance of incorporating multiple descriptors in studies where local roughness values play a crucial role in subsequent analyses. The spatial scales were found to have a smaller impact on rougher terrain, while interpolation methods had minimal influence on roughness maps derived from different descriptors

**Keywords:** roughness; digital elevation model; terrain; LiDAR; point cloud; remote sensing

## 1. Introduction

Terrain surface roughness is a key metric in the earth sciences that describes the complexity or variability of a terrain surface at a specific spatial scale. Its utility spans various applications, including geospatial analysis, simulation of Earth surface processes, and terrain classification [1]-[6]. Depending on specific application needs, calculations may involve determining either global or local terrain roughness. Especially when using point cloud data as the main data source, local terrain roughness is often calculated [7]-[11]. This is mainly due to the fine spatial resolution inherent in such data, enabling the detailed recording of local topographic surface characteristics and spatial changes. The acquisition of such data typically relies on Light Detection and Ranging (LiDAR), which has been widely used for characterizing the shape and form of land features [12].

The definition of terrain surface roughness frequently involves ambiguity. Roughness indicators typically depend on quantitative descriptions of alterations in specific terrain features, including factors like local relief, degree of folds, or the extent of local mutations. Commonly employed indicators include, but are not confined to standard deviation of residual elevation [9], root mean square height (RMSH) [11], standard deviation of slope [13]-[14], standard deviation of curvature [2], topographic ruggedness index [15], fractal dimension analysis [16], autocorrelation function [17] and geostatistical analysis [18]-[19],

As highlighted by Shepard et al. in 2001 [20], there was no standard method for quantitatively characterising surface roughness, a challenge that persists till now. The absence of a universally accepted or preferred method for estimating terrain surface roughness is largely due to the diverse range of applications and user requirements. In practical applications and research, individuals often choose a commonly used terrain surface descriptor based on personal preference, with limited consideration given to the validity of the local roughness map derived for a specific application. For rigorous applications, researchers evaluate multiple roughness descriptors to compare the results of interest to determine a more appropriate descriptor [7]-[11]. The comparison is usually based on simple visual inspection for the particular application under consideration [2], [9]. This approach requires more effort during the data processing stage.

Until now, there has been limited research dedicated to exploring quantitative correlations among metrics of terrain surface roughness, providing the motivation for this study. The research involves a comparative analysis of five frequently employed roughness descriptors, utilizing three lidar point cloud datasets that represent varying levels of terrain surface complexities.

## 2. Materials and Methods

### 2.1. Study data

This study considers three sets of airborne LiDAR data, each representing bare earth surfaces of approximately 350 metres by 350 metres. These terrain surfaces exhibit distinctive spatial variation characteristics: hilly rough, flat rough and flat smooth, with an average data spacing of 0.63-0.64 meters. The data are extracted from an extensive LiDAR dataset acquired by the National Airborne Laser Mapping Centre of the USA in a volcanic area in central Nevada [21]-[22].

Each set of point cloud data exhibits a noticeable global elevation trend. To minimise its impact on local surface roughness calculations and improve the visualization of surface spatial variability, the global trend within each set of point clouds is eliminated by subtracting the corresponding best-fitting plane. To prevent the display of negative elevation values in a detrended point cloud, detrended elevation values are subsequently translated upwards by an amount equal to the absolute value of the minimum elevation in the point cloud. This translation ensures that the minimum elevation within each detrended point cloud is set to zero, and enhances visual comparison of elevation ranges between different point clouds. Point cloud data with these adjusted elevations server as the study data for deriving local roughness maps in this study. These study data are shown in the left plots of Figure 1, with elevation represented by colour.

Additionally, the histograms in Figure 1 illustrate the distribution of elevation values, accompanied by the display of the range and the standard deviation of all elevation values. These two statistics provide insights into the overall roughness of the three terrain surfaces under consideration.

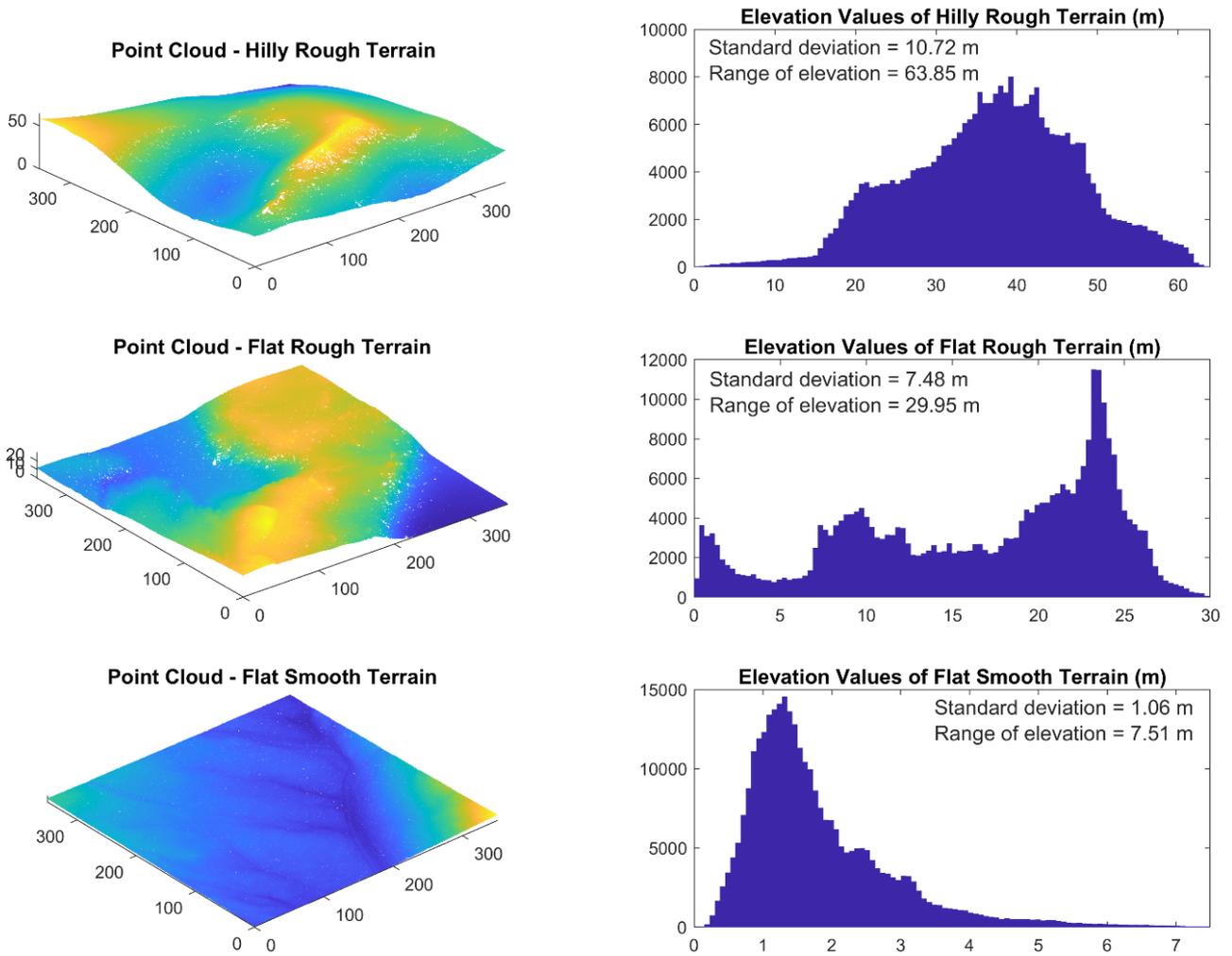

**Figure 1. Point cloud data (left) and elevation histograms (right).**

*2.2. Digital elevation model (DEM) maps*

Certain surface roughness descriptors, such as RMSH, are applicable to both unstructured scatter point cloud data and gridded (i.e., structured) data in the form of digital elevation models (DEMs). In contrast, many other descriptors like the standard deviation of curvature/slope are typically exclusive to gridded DEM maps.

To ensure a consistent comparison of various metrics for estimating roughness, gridded DEM maps are used to compute local surface roughness values in this study. To generate such maps, spatial interpolation is needed to convert scattered elevation data into a grid format. A range of interpolation methods is available, with simpler methods often preferred for high-density point cloud data [4].

The interpolation technique adopted in this study is natural neighbour interpolation, which identifies the nearest subset of known data points to a query grid location and assigns weights to them according to proportional areas to interpolate a value at the query location. It works well with irregularly distributed data points such as point cloud, and can produce smooth surfaces across scattered known data points while preserving their elevation values. This interpolation method does not require any user-defined parameters as input and therefore enhance consistency of generated DEM maps.

Considering the spatial resolutions (0.63 - 0.64 m) of the point cloud data used, a spatial grid resolution of 1 metre is employed for constructing DEM maps in this study. Figure 2 illustrates the DEM maps produced through the utilisation of detrended point cloud data, employing natural neighbour interpolation.

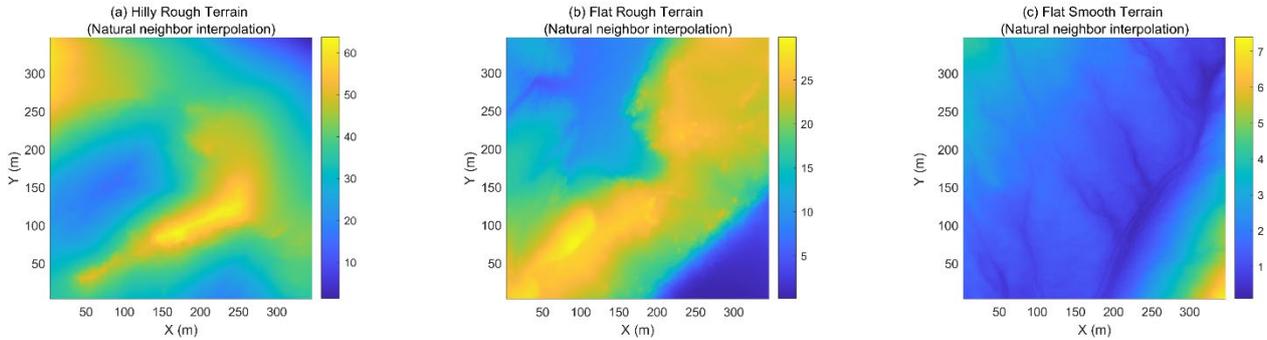

**Figure 2. DEM maps of the globally detrended point cloud data: (a) hilly rough terrain, (b) flat rough terrain and (c) flat smooth terrain, using triangulation with liner interpolation.**

### 2.3. Terrain surface roughness descriptors

This study investigates five commonly used descriptors for terrain roughness, with an elaboration of the computational procedures outlined in Section 2.3. These descriptors include RMSH, standard deviation of locally detrended residual elevations ($\sigma_{\text{LDRE}}$), standard deviation of residual topography ($\sigma_{\text{RT}}$), standard deviation of slope ($\sigma_{\text{slope}}$), and standard deviation of curvature ($\sigma_{\text{curvature}}$).

### 2.3.1. RMSH

RMSH serves as a commonly employed descriptor for measuring local surface roughness, especially when dealing with scattered elevation data [2], [23]. Its applicability extends to gridded elevation data in the form of a DEM. The definition of RMSH is provided in (1).

$$\text{RMSH} = \sqrt{\frac{\sum_{i=1}^{n}\left(Z_i - \overline{Z}\right)^2}{n-1}} \tag{1}$$

where $n$ denotes the number of selected data points; $Z_i$ represents the elevation value of the $i^{\text{th}}$ data point; $\overline{Z}$ denotes the mean elevation value of all ($n$) selected data points.

### 2.3.2. Standard deviation of locally detrended residual elevations

This approach involves linear detrending of local elevation data within a moving window. A best-fitting plane is utilized to derive residual elevation values, and the standard deviation of these residuals within the moving window is calculated to characterize local surface roughness.

### 2.3.3. Standard deviation of residual topography

Residual topography is characterized as the variance between the original DEM and the smoothed DEM [9]. In our investigation, the elevation value at a grid location in the smoothed DEM is established by averaging the elevation values of adjacent cells within a 5×5 moving window. Given that both the original and smoothed DEMs share the same spatial resolution, the residual topography is determined through arithmetic subtraction of the elevation values in corresponding cells between the two DEMs. The standard deviation of the residual topography is then computed as an indicator of terrain roughness.

### 2.3.4. Standard deviation of slope

Calculating the standard deviation of slope requires the computation of slope values. Slope represents the rate of change of terrain elevations and is expressed in (2).

$$\text{slope} = \tan^{-1}\left(\sqrt{\left(\frac{dz}{dx}\right)^2 + \left(\frac{dz}{dy}\right)^2}\right) \tag{2}$$

where $dz/dx$ and $dz/dy$ denote the rate of change in the $x$ and the $y$ directions, respectively, for the cell under consideration.

In the context of DEM maps, slope is often determined using elevation values in a 3×3 moving window, as shown in (3). The calculations of $dz/dx$ and $dz/dy$ for the central cell ($Z_5$) are determined through (4) and (5), respectively, where $L$ represents the cell size. In cases where neighbouring cells (such as those at the edge of a DEM) do not have elevation data, it is assumed that these cells adopt the elevation value of the central cell. This approach is valuable for cells positioned at the DEM raster's edge, ensuring consistency in spatial extent between the slope map and the DEM map.

$$\text{Moving Window} = \begin{bmatrix} Z_1 & Z_2 & Z_3 \\ Z_4 & Z_5 & Z_6 \\ Z_7 & Z_8 & Z_9 \end{bmatrix} \tag{3}$$

$$\frac{dz}{dx} = \frac{\left[\left(Z_3 + 2Z_6 + Z_9\right) - \left(Z_1 + 2Z_4 + Z_7\right)\right]}{8L} \tag{4}$$

$$\frac{dz}{dx} = \frac{\left[\left(Z_7 + 2Z_8 + Z_9\right) - \left(Z_1 + 2Z_2 + Z_3\right)\right]}{8L} \tag{5}$$

### 2.3.5.  Standard deviation of curvature

Curvature is determined by computing the second derivative of a DEM map, using the same moving window as empolyed for slope calculation. Various methods are avlaible for calculating curvature. This study employs the approach presented by Zevenbergen and Thorne [24]-[25], outlined in equations (6) to (8). Similar to the procedure for slope calculation, in the presence of non-value cells in the neighbourhood, it is assumed that these cells adopt the elevation value of the central cell. Following the derivation of the curvature map, the standard deviation of curvature is computed to characterize terrain roughness.

$$\text{curvature} = 2E + 2D \tag{6}$$

where $D$ and $E$ are given in (7) and (8), respectively, where relevant elevation values are those specified in the moving window shown in (3).

$$D = \frac{\left[\left(Z_4 + Z_6\right)/2 - Z_5\right]}{L^2} \tag{7}$$

$$E = \frac{\left[\left(Z_2 + Z_8\right)/2 - Z_5\right]}{L^2} \tag{8}$$

### 2.4. *Correlation between roughness maps*

To assess the correlation between roughness maps, this study computes the correlation coefficient ($r$) of the pixel values of two maps compared, using (9), which is widely used for assessing correlation between images. A correlation coefficient of 1 represents that the pixel values of two compared maps are perfectly matched at all pixel locations. A correlation coefficient of 0 suggests that the pixel values in one map are randomly different from the corresponding pixel values in the other map.

$$r = \frac{\sum_m \sum_n \left(A_{mn} - \overline{A}\right)\left(B_{mn} - \overline{B}\right)}{\sqrt{\left(\sum_m \sum_n \left(A_{mn} - \overline{A}\right)^2\right)\left(\sum_m \sum_n \left(B_{mn} - \overline{B}\right)^2\right)}} \tag{9}$$

where $A$ and $B$ denote the pixel values of two compared roughness maps, respectively; the subscripts $m$ and $n$ refer to the pixel location in the maps; $\overline{A}$ and $\overline{B}$ denote the corresponding mean value.

*2.5. Spatial scale for roughness calculation*

One of the main challenges in studying roughness is its highly scale-dependent nature [26]. Local terrain surface roughness is often determined at a particular user-defined spatial scale. This is typically achieved through non-overlapping moving windows in the context of DEM maps. However, the extent to which the spatial scale influences the correlation between roughness maps characterized by different descriptors remains unclear. Consequently, this study explores various non-overlapping moving window sizes, including 3×3, 5×5, 7×7, 9×9, and 11×11 moving windows, to assess their potential impact.

## 3. Results

*3.1. Roughness maps by different descriptors*

Figure 3 depicts local terrain roughness maps for the five roughness descriptors, utilising a spatial scale of 5×5 cells (a commonly used spatial scale in practice) as an illustrative example. In Figure 3, each column in the plots represents a different roughness descriptor while each row corresponds to one of the three considered terrain surfaces. To enhance visual comparison, the roughness values shown in Figure 3 were normalized to a range of 0 to 1.

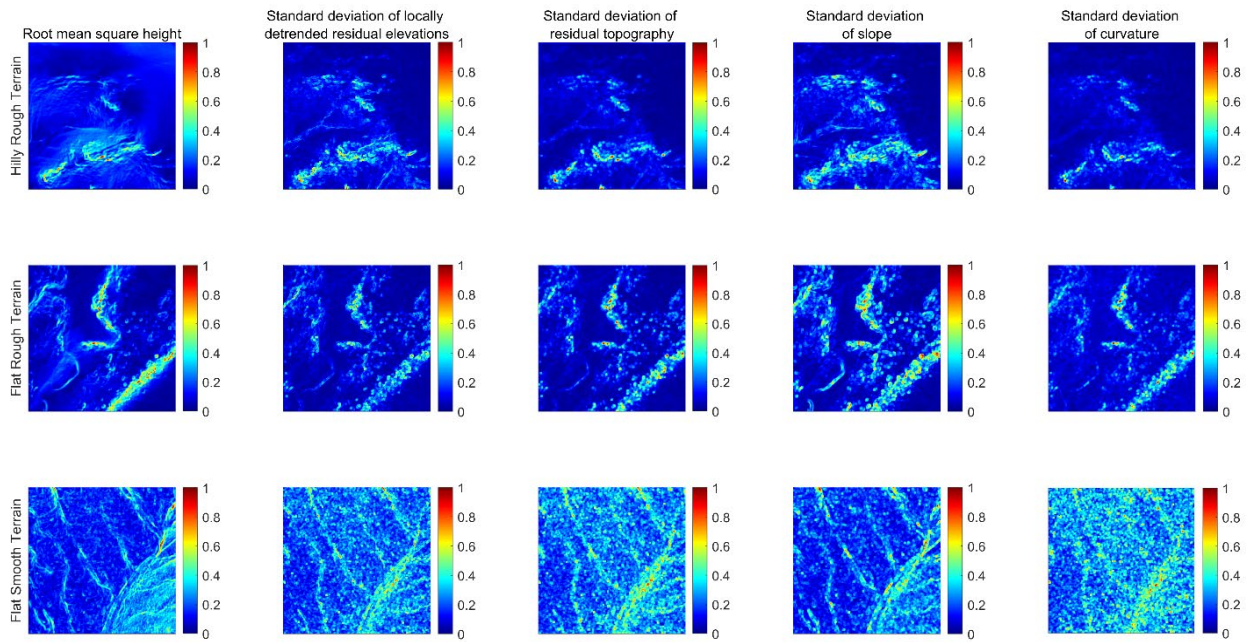

**Figure 3. Local roughness maps for the three terrain surfaces, derived using the DEM maps in Figure 2.**

Figure 3 shows that terrain roughness produced by most roughness descriptors exhibited similar global patterns, justifying their widespread use in the literature. However, the terrain roughness maps produced by RMSH appeared to be more distinct from the others. In terms of local variations in the distributions of roughness values among different descriptors, certain descriptors (such as standard deviation of residual topography and standard deviation of curvature) exhibited similar local distributions, while others (such as RMSH and standard deviation of slope) showed notably differences.

These distinctions between descriptors imply that the choice of roughness descriptors can impact the results of subsequent analyses, particularly in quantitative studies reliant on local roughness values. As such, it is advisable to carry out a sensitivity analysis using multiple roughness descriptors in quantitative studies where local roughness values are the critical inputs.

### 3.2. Quantified correlations between roughness maps

The correlation coefficient $r$ for each pair of compared roughness maps was calculated using (8) for the three terrain surfaces considered. With five roughness descriptors in consideration, a total of 10 pairs were formed for comparison. The correlation coefficient values are visually presented in the radar charts depicted in Figure 4. In these radar charts, each vertex corresponds to a roughness descriptor, and markers along each axis from the radar centre (where $r = 0$) to the vertex (where $r = 1$) represent the correlation coefficient values between other descriptors and the one located at the vertex.

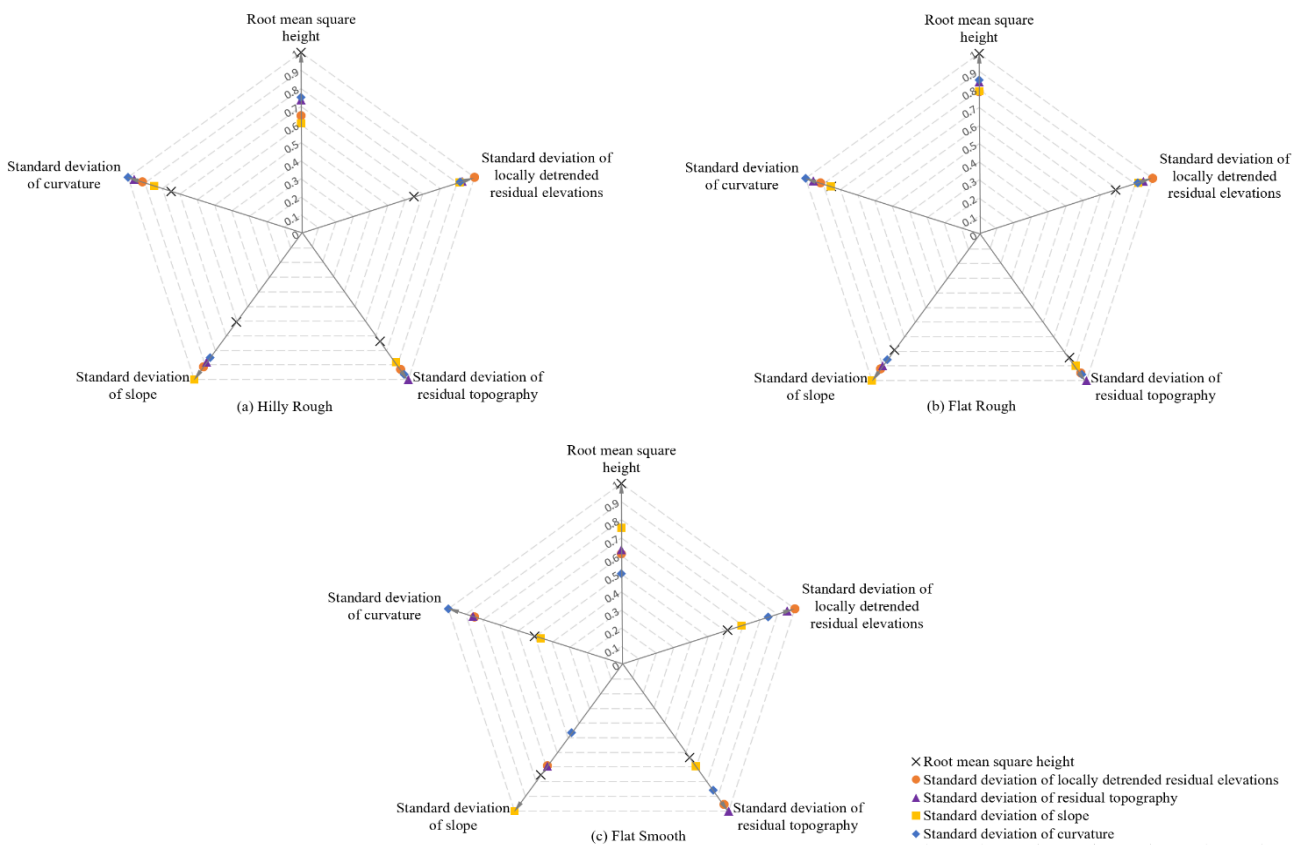

**Figure 4.** **Correlation coefficient values for paired roughness maps. In the charts, each vertex corresponds to a roughness descriptor, and markers along each axis from the radar centre to the vertex represent the correlation coefficient values between other descriptors and the one located at the vertex.**

In the case of rough terrain surfaces (both hilly rough and flat rough), large correlations were identified between roughness maps derived using the considered descriptors, with the exception of RMSH, as shown in Figure 4. Notably, the largest correlation values (approximately 0.964 for hilly rough terrain and 0.960 for flat rough terrain) were observed for the pair $\sigma_{\mathrm{RT}}$ and $\sigma_{\mathrm{curvature}}$, aligning with the global patterns of the roughness maps shown in Figure 3 for these two descriptors. However, for the flat smooth terrain, weak correlations between different roughness descriptors were observed. This suggests that the choice of roughness descriptors imposes a greater impact on the estimated roughness of terrains that are less rough.

In Figure 4, the correlation coefficient values for the flat rough terrain were found to surpass those for the hilly rough terrain. This was likely attributable to distinct characteristics in the spatial variations between the two terrain surfaces. For the hilly rough terrain, its spatial variation comprised a strong signal spatial variation (as indicated by "hilly"), alongside a noisy spatial variation. In contrast, the flat rough terrain was predominately featured by noisy spatial variations with minimal signal spatial variations (as indicated by "flat"). The characterisation of the signal component likely varied with algorithms (i.e., roughness descriptors in this study), resulting in greater differences between roughness maps produced across different roughness descriptors. Generally, the similarity between roughness maps generated by different descriptors was influenced by the magnitude and type of spatial variations. Greater similarity was observed for rougher terrain surfaces, especially those with noisy spatial variations.

The correlation between RMSH and any of the other four roughness descriptors was found to be relatively modest, as depicted in Figure 4. This observation was also supported by the roughness maps presented in Figure 3. Notably, in Figure 3, the roughness values of RMSH exhibited greater spatial coherence compared to those derived from the other descriptors. This phenomenon can likely be attributed to the presence of local elevation trends, resulting in stronger local spatial autocorrelation in the RMSH maps. This explanation is supported by the roughness map of $\sigma_{\mathrm{LDRE}}$, which shares the same algorithm as RMSH but uses locally detrended elevation values.

*3.3. Impact of spatial scales*

The influence of spatial scales, represented by the moving window size used, on the correlations between roughness maps generated by different roughness descriptors are illustrated in Figure 5. A mixed behaviour was observed. When comparing two paired roughness descriptors, the correlation between them could either increase or decrease with a larger window size.

Notably, the terrain complexities exerted a great impact, as evidenced by the results presented in Figure 5. The changes of correlation coefficient values with varying spatial scales were significant for the flat smooth terrain. However, for the flat rough terrain, such changes were relatively small for most paired descriptors. This seems to suggest that the impact of the window size on correlations was smaller for the rougher terrain surfaces, especially those characterised by greater noisy spatial variations, such as the flat rough terrain.

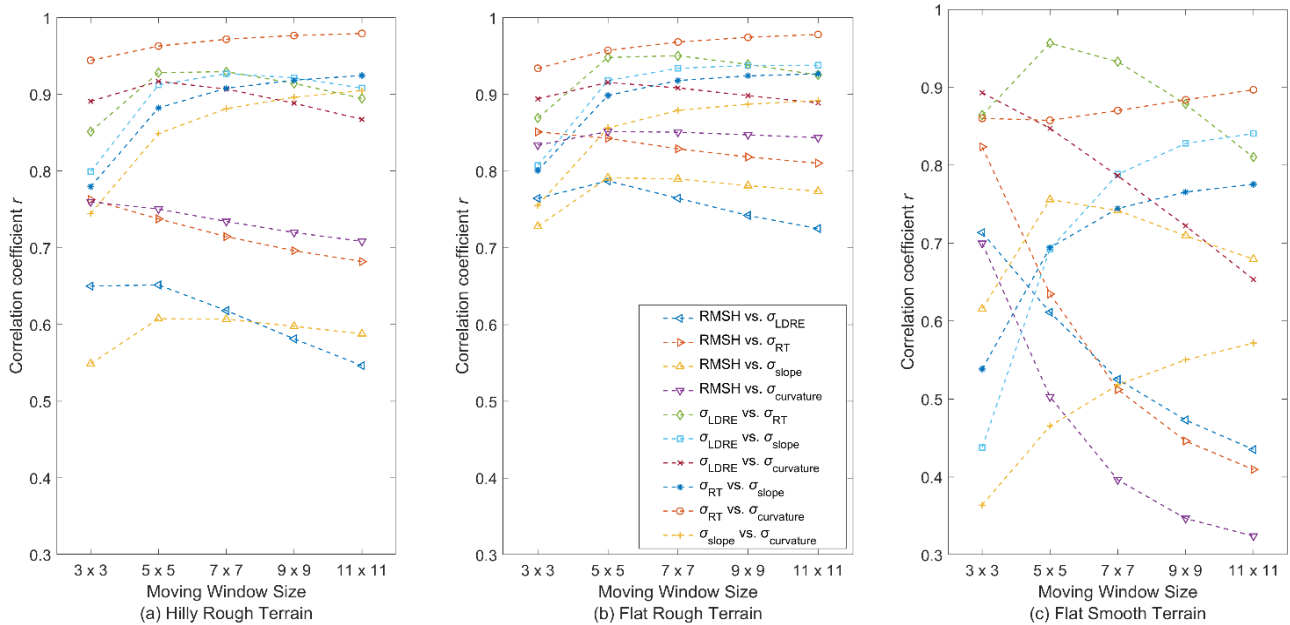

**Figure 5. Correlation between paired roughness maps under different window sizes: (a) hilly rough terrain, (b) flat rough terrain, (c) flat smooth terrain.**

## 4. Discussions

To understand the potential impact of interpolation methods on roughness maps generated from different roughness descriptors, this study also considered another two simple yet commonly used interpolation techniques for rasterising point cloud data, including nearest neighbour (i.e., assigning the elevation value of the nearest known data point to a query grid location) and triangulation with linear interpolation (i.e., using the Delaunay triangulation to form triangles, and determining the elevation at a query grid location through linear interpolation using the triangle's three vertices [27]). Other more advanced interpolation techniques are not considered because the impact of interpolation techniques on DEM accuracy is typically minimal when applied to highly dense data such as LiDAR point clouds. The impact of interpolation methods on roughness maps generated from various roughness descriptors is illustrated in Figure 6. Large correlation coefficient values suggest neglectable differences between roughness maps from DEMs through triangulation with liner interpolation and natural neighbour interpolation. In comparison, roughness maps derived from DEMs through nearest neighbour were slightly more distinct from those from DEMs through triangulation with liner interpolation, especially for the flat smooth terrain. These observations suggest that the influence of interpolation methods on roughness maps derived from different roughness descriptors is minimal, likely attributed to the high-density point cloud data used.

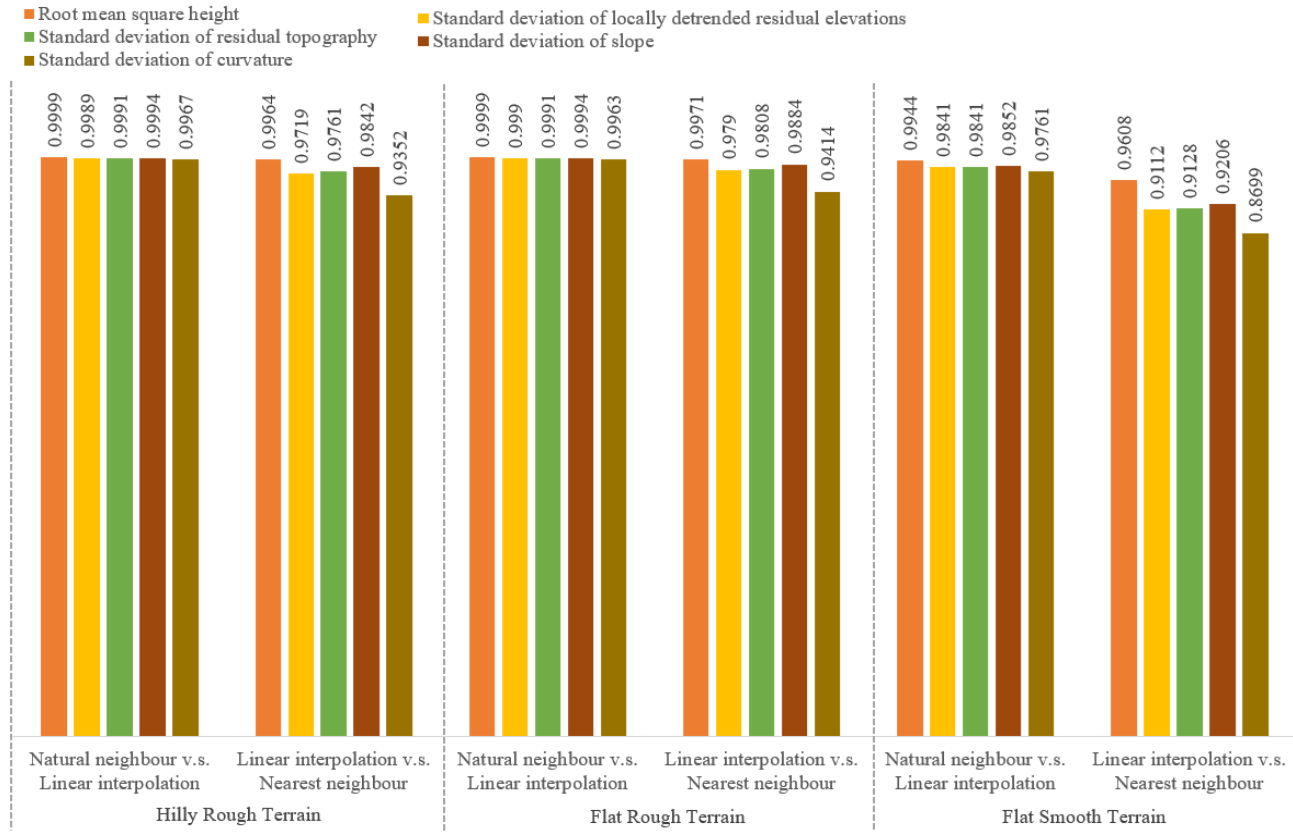

**Figure 6. Correlation coefficient values for compared roughness maps from different interpolation methods.**

In this study, it was observed that different algorithms yielded diverse roughness values. Therefore, the selection of an algorithm should align with the specific goals of a geospatial analysis. When integrating terrain roughness information with other geospatial data layers such as slope and aspect, careful consideration is essential to ensure compatibility and meaningful integration. For instance, in a geospatial analysis task combining surface roughness with slope data, it may be more appropriate to utilize the standard deviation of slope as the roughness descriptor for compatibility and meaningful integration.

While this study focuses on differences between roughness maps generated by various descriptors, it does not look into the accuracy of terrain roughness values. It is evident that roughness maps are susceptible to various uncertainties. For example, the accuracy and resolution of point cloud data can impact the accuracy of DEM maps, subsequently affecting the accuracy of calculated roughness values. Coarse-resolution or inaccurate DEM maps may fail to capture subtle terrain variations, leading to underestimation or overestimation of roughness. Currently, there are limited established methods for quantifying these uncertainties, and future research in this regard would enhance roughness accuracy and contribute to and more informed decision-making.

Machine learning algorithms have gained widespread use in geoscience. Future research could explore the application of machine learning algorithms for predicting and classifying terrain surface roughness. Training models to automatically identify roughness characteristics from LiDAR data has the potential to reduce the manual effort required in the analysis.

In this study, local roughness values are computed using LiDAR-derived DEM maps. As potential future work, a comparison could be made with methods directly applied to scattered point cloud data, such as RMSH of data points, the standard deviation of orthogonal point-to-plane distances [28], and the degree of dispersion of normal vectors among adjacent data points.

## 5. Conclusions

In this study, the roughness maps of three terrain surfaces of distinct spatial variations, quantified by five commonly used roughness descriptors, were compared. The following findings are found.

1. Local roughness maps from the considered descriptors exhibited similar global patterns across varying spatial complexities, demonstrating their effectiveness in characterizing terrain surface roughness.

2. Similarity between roughness maps generated by different descriptors was influenced by the magnitude and type of spatial variations, with greater similarity observed for rougher terrain surfaces, particularly those with noisy spatial variations.

3. Variations in local distributions of roughness values were noted among descriptors, highlighting the significance of considering multiple roughness descriptors in cases where local roughness values serve as inputs for subsequent analyses, especially for the widely used RMSH, which showed small correlations with the other descriptors.

4. Investigation of the spatial scale of roughness maps revealed mixed effects on correlations between two roughness descriptors, with a smaller impact on correlations for rougher terrain surfaces, especially those with greater noisy spatial variations like the flat rough terrain.

5. Minimal influence of interpolation methods on roughness maps derived from different descriptors was observed, likely due to the high density of the point cloud data used.

**Acknowledgments (All sources of funding of the study must be disclosed)**

The author is grateful for the financial support from Xi'an Jiaotong–Liverpool University, which includes the Research Enhancement Fund (grant number REF-21-01-003).